\documentclass[10pt,twocolumn,letterpaper]{article}

\usepackage{cvpr}              

\usepackage[dvipsnames]{xcolor}

\definecolor{cvprblue}{rgb}{0.21,0.49,0.74}
\usepackage[pagebackref,breaklinks,colorlinks,citecolor=cvprblue]{hyperref}

\def\confName{CVPR}
\def\confYear{2024}
\def\paperID{*****} 

\title{ImagineMap: Enhanced HD Map Construction with SD Maps}

\author{Yishen Ji \\
\and 
Zhiqi Li \\
Nanjing University \\[2ex]
December 22, 2024 
\and 
Tong Lu \\
}

\begin{document}
\maketitle
\begin{abstract}

Track Mapless demands models to process multi-view images and Standard-Definition (SD) maps, outputting lane and traffic element perceptions along with their topological relationships. We propose a novel architecture that integrates SD map priors to improve lane line and area detection performance. Inspired by TopoMLP, our model employs a two-stage structure: perception and reasoning. The downstream topology head uses the output from the upstream detection head, meaning accuracy improvements in detection significantly boost downstream performance. 

\end{abstract}    
\section{Introduction}
\label{sec:intro}


Driving without High-Definition (HD) maps requires a vehicle to possess a high level of active scene understanding capability. This challenge aims to explore the limits of scene reasoning. Neural networks need to process multi-view images and Standard-Definition (SD) maps, and output perception results of lanes and traffic elements, while simultaneously providing topological relationships among lanes and between lanes and traffic elements. This means that the neural network must not only identify lanes and various traffic elements on the road but also understand the interrelationships and positional relationships between them, thereby enhancing the driving capability of autonomous systems in the absence of HD map support.

In this work, we propose a novel architecture that integrates SD map priors into the model, thereby enhancing the performance of lane line and area detection. 
Our overall model design is inspired by TopoMLP \cite{wu2023topomlp} and employs a two-stage structure, perception and reasoning. The downstream topology head takes the output of the upstream detection head as its input, meaning that improvements in the accuracy of the detection head will significantly boost the performance of the downstream head. 
We utilize DETR for constructing the lane segment detection head and deformable DETR for constructing the traffic element head. 
\section{Method}
\label{sec:method}

\subsection{Lanesegment Detection}

\begin{figure}[t]
  \centering
   \includegraphics[width=1\linewidth]{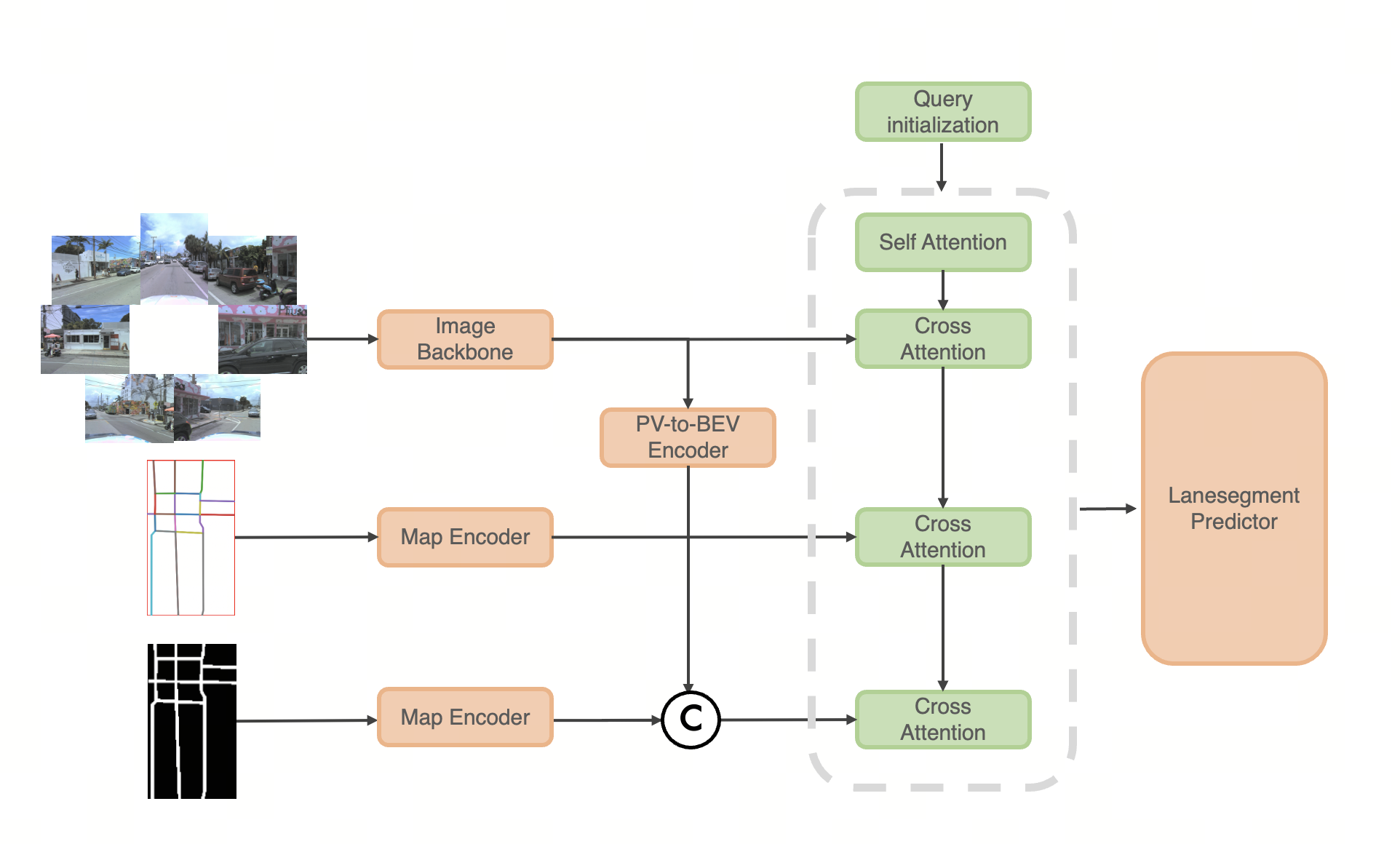}
   \caption{The structure for incorporating SD map information into detection. Map encoder extracts SD map features to interact with lane queries via cross-attention. Additionally, we rasterize SD map to generate a mask, whose features are extracted by a small map backbone and concatenated with BEV features from BEV encoder for further interaction with lane queries.}
   \label{fig:lane}
\end{figure}

Our lane detection head is based on LaneSegNet\cite{li2023lanesegnet} and adopts the DETR\cite{carion2020end} structure. Each lane segment includes three lines, a centerline and its corresponding left and right boundaries, and each line is represented by a series of evenly spaced three-dimensional points. The number of sample points for each line is $ N_p $. We adopt instance query to represent the lane segment. 

By utilizing the position and posture information of ego vehicle, we obtain the SD map (vectorized) within the BEV range. 
To fully harness the information provided by the SD map, we incorporate it into our detection process in two ways.
First, we employ SMERF\cite{luo2023augmenting}, a powerful transformer-based encoder, to extract the SD map features. These are then used as queries to interact with the lane queries through cross-attention.
Next, we grid the SD map to generate a mask and extract its features through a small network(ResNet-18). These features are concatenated with the BEV features constructed by BEVFormer\cite{li2022bevformer} and again interact with the lane queries through cross-attention. The specific structure is illustrated in Figure \ref{fig:lane}.

To further leverage semantic and geometric information, we introduce auxiliary foreground segmentation on bird’s eye view (BEV). 
The entire lane segment area (bounded by the left and right lane lines) is considered as the segment mask range.

We use Focal loss for the class head and L1 loss for the lane head. We introduced a lane line type classification loss using Cross Entropy Loss. 
BEV seg loss is composed of mask loss(Cross Entropy Loss) and dice loss.
Thus, the overall lanesegment loss function can be formulated as:

\begin{align}
L_{ls} &= \lambda_{cls}L_{cls} + \lambda_{reg}L_{reg} + \lambda_{type}L_{type} \notag \\
       &\quad + \lambda_{mask}L_{mask} + \lambda_{dice}L_{dice}
\label{eq:l_ls}
\end{align}

where $ \lambda_{cls}, \lambda_{reg},\lambda_{type},\lambda_{mask},\lambda_{dice}$ is set to 1.5, 0.0025, 0.1, 3.0 and 3.0 respectively.

\subsection{Area Detection}

We implemented the area detection task based on MapTr\cite{liao2022maptr} structure.

Similar to the way we utilize the sd map prior in lane detection task, we pass the sd map through a separate map encoder, and then perform cross-attention with the query on the features.

Auxiliary BEV segmentation supervision is also adopted in this task.
But unlike in the lane detection task where the entire lane segment is considered for segmentation, the area mask is only taken along the instance boundary.

The area detection head is supervised by $L_a$, which is composed of a classification loss $ L_{cls}$ (Focal Loss), a regression loss $ L_{reg}$ (L1 Loss), a direction loss $ L_{dir}$ (Points Direction Cosine Loss) and auxiliary seg loss $ L_{seg}$ (Cross Entropy Loss):

\begin{equation}
L_a = \lambda_{cls}L_{cls} + \lambda_{reg}L_{reg}  +  \lambda_{dir}L_{dir} + \lambda_{seg}L_{seg}
\label{eq:l_a}
\end{equation}

where $ \lambda_{cls}, \lambda_{reg},\lambda_{dir},\lambda_{seg}$ is set to 1.5, 0.0025, 0.005, 10 respectively. 

\subsection{Traffic Element Detection}

Our traffic element detection head follows the head design in Deformable DETR\cite{zhu2020deformable} to predict bounding boxes and classification scores. The detection head takes the front-view image features as input. To improve the accuracy of the detection head, we use Co-DETR\cite{zong2023detrs}, a powerful object detection model, to generate proposals.
Co-DETR is a novel collaborative hybrid assignments training scheme for more efficient and effective DETR-based detectors. This scheme enhances the learning ability of the encoder in end-to-end detectors through training multiple parallel auxiliary heads supervised by one-to-many label assignments. It also improves the attention learning of the decoder by conducting extra customized positive queries, extracting the positive coordinates from these auxiliary heads.

The traffic elements detection head is supervised by $L_{te}$,
which decomposed into a classification loss $ L_{cls}$ (Focal
Loss), a regression loss $ L_{reg}$ , and an IoU loss $ L_{iou}$ (GIoU):

\begin{equation}
L_{te} = \lambda_{cls}L_{cls} + \lambda_{reg}L_{reg}  +  \lambda_{iou}L_{iou}
\label{eq:l_te}
\end{equation}

where $ \lambda_{cls}, \lambda_{reg},\lambda_{iou}$ is set to 1.0, 2.5, 1.0 respectively.

\subsection{Topology Lane-lane}

The lane-lane topology reasoning branch aims to predict the connection relationship between lanes. To incorporate the distinct lane information, we integrate the predicted lane points into the lane query features. We use a MLP to embed the lane coordinates and then add them to the decoded lane query features.  
The lane-lane topology head is supervised by $L_{ll}$ (Focal Loss), with a weight of 5.

\subsection{Topology Lane-Traffic}

Lane-traffic topology head is also an MLP structure but simultaneously takes inputs from both the lane head and the traffic head.
The lane-traffic topology head is supervised by $L_{lt}$ (Focal Loss), with a weight of 5.


\section{Experiments}

\subsection{Dataset and Metrics}

OpenLane-V2 is the first dataset on topology reasoning for traffic scene structure. 
This dataset utilizes multi-view images as input and encompasses a range of tasks, which include the perception of lane lines and traffic elements, the detection of specific areas, and the prediction of topological relationships between lane lines as well as between lane lines and traffic elements. 
OpenLane-V2 is constructed on the base of two existing datasets: Argoverse 2\cite{wilson2023argoverse} (subset A) and nuScenes\cite{caesar2020nuscenes} (subset B). 
In scenarios involving mapless tracks, we employ the subset A dataset.

For subset A, the resolution of input images is 2048 × 1550, except for the front-view image, whose resolution is 1550 × 2048.
The number of scene segment for Train set, Val set and Test set is 700, 150 and 150. 
And the annotation range for the dataset is +-50 meters in x-axis and +-25 meters in y-axis(for the vehicle coordinate system)

The evaluation system utilizes OpenLane-V2 UniScore (OLUS) as the final metric, which is the average of various metrics covering different aspects of the primary task.The calculation for OLUS is as follows:

$$ OLUS = \frac{1}{5} [DET_l + DET_a + DET_t + \sqrt{TOP_{ll}} + \sqrt{TOP_{lt}} ] $$

$DET_l$: lanesegment prediction with Frechet(for centerline) and Chamfer(for left and right laneline) based mAP;
$DET_a$: area prediction with Chamfer based mAP;
$DET_t$: traffic element prediction IOU based mAP;
$TOP_{ll}$: mAP on topology among lane segments;
$TOP_{lt}$: mAP on topology between lane segments and traffic elements;

It is important to note that this year's metrics differ from that of last year. The task of lane detection now includes the centerline and the left and right lanelines. Moreover, there's a new prediction task: area, namely pedestrian crossings and road boundaries, are regarded as undirected curves, which are closed or intersected with the boundaries of the BEV range.

\subsection{Experimental Details}

Initially, the front-view image is cropped to a size of $ 1550 \times 1550 $. Subsequently, all images are scaled down by a factor of 0.75 as the input to backbone to achieve a balance between performance and training speed.

We utilize the powerful EVA02 \cite{fang2023eva} as our backbone to extract multi-view image features, which are then used in the lane segment (ls) head and traffic element (te) head (te head only uses features from the front-view image). We build the bird's eye view (BEV) features using BEVFormer\cite{li2022bevformer}, and the number of encoder layer is set as 6. Increasing the size of BEV features can also improve the performance of lane detection and area detection. We increased it from $ 200\times100$ to $ 400\times200$. However, this will correspondingly require more computational resources. 

We adopt the AdamW optimizer and a cosine annealing schedule with an initial learning rate of 4e-4. Our proposed method is trained for 24 epochs with a batch size of 1, using 16 NVIDIA Tesla A100 GPUs. We also employ layer decay to prevent overfitting and accelerate convergence.

\subsection{Lane Detection}

We found that the detection performance is optimal when the number of sampling points 
$ N_p $ for each lane query matches the number of points in the annotations. We set $N_P= 11$. And number of instance query is 300.

We tested the impact of different auxiliary methods on lane detection performance on the test set, as shown in Table \ref{tab: lane ablation}.

\begin{table}[h]
    \centering
    \begin{tabular}{lcc}
        \toprule
        Method & mAP (\%) \\
        \midrule
        Base & 32.22 \\
        + EVA02 + Layer Decay & 35.03 \\
        + BEV Scale Up & 37.66 \\
        + Scale 0.75 & 39.08 \\
        + Aux BEV Seg Sup & 42.95 \\
        \bottomrule
    \end{tabular}
    \caption{Ablation of lanesegment detection on OpenLane-V2 test set.}
    \label{tab: lane ablation}
\end{table}

\subsection{Area Detection}

Abalation on area detection is shown in Table \ref{tab: area ablation}. 

Fine tune can further improve the performance of area head. We freeze the other parts of the model and train the area head for another 24 epochs using a smaller learning rate.

\begin{table}[h]
    \centering
    \begin{tabular}{lcc}
        \toprule
        Method & mAP (\%) \\
        \midrule
        Base & 20.52 \\
        + EVA02 Layer Decay & 23.46 \\
        + BEV Scale Up & 26.59 \\
        + SD Map Query & 28.00 \\
        + Scale 0.75 & 30.76 \\
        + Aux BEV Seg Sup & 32.00 \\
        + Fine Tune & 34.72 \\
        \bottomrule
    \end{tabular}
    \caption{Ablation of area detection on OpenLane-V2 test set.}
    \label{tab: area ablation}
\end{table}

\subsection{Traffic Element Detection}

Introducing proposals generated by Co-DETR significantly improved the traffic element detection performance. To apply Co-DETR on OpenLane-V2 dataset, we first get all of the front-view images from multi-view images (and corresponding ground truth), then reformat them according to the COCO\cite{lin2014microsoft} dataset format. Input images is cropped into $1550\times1550$ for training effciency.  
We tested the performance of two strong backbones. Using Swin-L \cite{liu2021swin} to generate proposals, TE detection performance achieved 79.15. In contrast, using ViT-L to generate proposals, TE detection performance improved to 82.15, which is shown in Table \ref{tab: te ablation}.

\begin{table}[h]
    \centering
    \begin{tabular}{lcc}
        \toprule
        Method & mAP (\%) \\
        \midrule
        Base & 67.47 \\
        + proposal & 79.15 \\
        + vit-L & 82.15 \\
        \bottomrule
    \end{tabular}
    \caption{Ablation of traffic element detection on OpenLane-V2 test set.}
    \label{tab: te ablation}
\end{table}

\subsection{Topology}

As shown in Table \ref{tab: ll ablation} and Table \ref{tab: lt ablation}, methods that improve the performance of upstream detection heads can effectively enhance the reasoning performance of the topology heads.

\begin{table}[h]
    \centering
    \begin{tabular}{lcc}
        \toprule
        Method & mAP (\%) \\
        \midrule
        Base & 29.99 \\
        + EVA02 Layer Decay & 31.57 \\
        + lane query300 & 36.48 \\ 
        \bottomrule
    \end{tabular}
    \caption{Ablation of lane-lane topology on OpenLane-V2 test set.}
    \label{tab: ll ablation}
\end{table}

\begin{table}[h]
    \centering
    \begin{tabular}{lcc}
        \toprule
        Method & mAP (\%) \\
        \midrule
        Base & 32.98 \\
        + TE proposal & 37.56 \\
        + EVA02 Layer Decay & 40.64 \\
        + lane query300 & 41.91 \\
        \bottomrule
    \end{tabular}
    \caption{Ablation of lane-traffic topology on OpenLane-V2 test set.}
    \label{tab: lt ablation}
\end{table}

{
    \small
    \bibliographystyle{ieeenat_fullname}
    \bibliography{main}
}


\end{document}